\documentclass{ecai}
\usepackage{times}
\usepackage{graphicx}
\usepackage{latexsym}

\usepackage{cite}
\usepackage{amsmath,amssymb,amsfonts}
\usepackage{algorithm}
\usepackage[noend]{algpseudocode}

\usepackage{graphicx}
\usepackage{textcomp}
\usepackage{xcolor}

\usepackage[utf8]{inputenc}
\usepackage{cite}
\usepackage{amsmath,amssymb,amsfonts}

\usepackage{graphicx}
\usepackage{textcomp}
\usepackage[center]{subfigure}
\usepackage{multirow}
\usepackage{bm}
\usepackage{siunitx}
\usepackage[
             pdfstartview=FitH,
             plainpages=false,
             bookmarksopen=true,
             bookmarksnumbered=true,
             colorlinks=true,
             bookmarkstype=toc]{hyperref} 

\usepackage{xcolor} 

\graphicspath{{img/},{logs/}}

\def\BibTeX{{\rm B\kern-.05em{\sc i\kern-.025em b}\kern-.08em
    T\kern-.1667em\lower.7ex\hbox{E}\kern-.125emX}}

\newcommand\blfootnote[1]{%
  \begingroup
  \renewcommand\thefootnote{}\footnote{#1}%
  \addtocounter{footnote}{-1}%
  \endgroup
}

\begin{document}

\title{Robot self/other distinction: active inference meets neural networks learning in a mirror \blfootnote{Accepted at European Conference on Artificial Intelligence (ECAI 2020)}}

\author{Pablo Lanillos\institute{Donders Institute for Brain, Cognition and Behaviour, the Netherlands.} \and Jordi Pages\institute{PAL robotics, Pujades 77-79, 4-4 08005 Barcelona, Spain.} \and  Gordon Cheng\institute{Institute for Cognitive Systems, Technical University of Munich, Arcisstrasse 21, 80333 Munich, Germany.}}

\maketitle
\bibliographystyle{ecai}

\begin{abstract}
Self/other distinction and self-recognition are important skills for interacting with the world, as it allows humans to differentiate own actions from others and be self-aware. However, only a selected group of animals, mainly high order mammals such as humans, has passed the mirror test, a behavioural experiment proposed to assess self-recognition abilities. In this paper, we describe self-recognition as a process that is built on top of body perception unconscious mechanisms. We present an algorithm that enables a robot to perform non-appearance self-recognition on a mirror and distinguish its simple actions from other entities, by answering the following question: am I generating these sensations? The algorithm combines active inference, a theoretical model of perception and action in the brain, with neural network learning. The robot learns the relation between its actions and its body with the effect produced in the visual field and its body sensors. The prediction error generated between the models and the real observations during the interaction is used to infer the body configuration through free energy minimization and to accumulate evidence for recognizing its body. Experimental results on a humanoid robot show the reliability of the algorithm for different initial conditions, such as mirror recognition in any perspective, robot-robot distinction and human-robot differentiation.

\end{abstract}


\section{Introduction}
\label{sec:main}
\begin{quote}
    \textit{``Our bodies are our gardens, to the which our wills are gardeners." Othello 1603. William Shakespeare.}
\end{quote}

Humans perceive their body in a flexible manner \cite{botvinick1998rubber} but maintaining their identity \cite{gallagher2000philosophical}. Around the age of two, the majority of infants are able to pass the \textit{self-recognition mirror test}\cite{rochat2003five}, like elephants, dolphins and some non-human primates \cite{anderson2011primates}. This experiment has been argued to be hardly intertwined with self-awareness and social understanding \cite{gold2009using} and thus consciousness. Recently, a cleaner wrasse fish reopened the debate about the suitability of this test for evaluating self-awareness \cite{kohda2019if}. As reported by the authors, it was the first fish to pass the mirror test, waving the notion of consciousness by suggesting different evolutionary paths and showing a broad spectrum of self-awareness. 

Self-recognition is commonly understood as a reflective process, i.e. you are aware and able to think about it. However, it can be treated in two different levels \cite{synofzik2013experience}: the sensorimotor \cite{mitchell1997kinesthetic} and the cognitive \cite{anderson2011primates}. Regarding the first level, we hypothesize that the mirror test can be partially passed without being self-conscious, opening the door for artificial agents to self-recognize and to acquire the skill of self/other distinction. Actually, underneath self-recognition, we need body perception and action processes, and the first step to achieve it is contingency-learning \cite{lanillos2016yielding}. According to \cite{blakemore2001perception}, humans can learn and predict the consequence of actions and these forward models are used to determine the source of the sensory events. Thus, in this work we extrapolated this concept to robots. The agent should be able to answer the question ``is this my body?" by answering ``am I generating those effects in the world?".

We present an algorithm that enabled a robot with non-appearance self-recognition capabilities, by addressing the relation between forward models prediction error and its use for determining the source of the sensory cues. In other words, the artificial agent can self-recognize through accumulating evidence of producing effects in the world by self-generated actions. For that purpose, the robot needs to learn the expected sensory effect of: the changes in the visual field when it moves the body and where its body (e.g. hand) will appear in the visual field when it is in front of the mirror.

Previous works on artificial self-recognition, such as \cite{lanillos2016yielding,gold2009using,stoytchev2011self}, have been hardly criticized \cite{anderson2015mirror}: ``Merely simulating certain features of self-recognition through training/programming does not mean that the underlying mechanisms are the same, similar, or even remotely related". Actually, we agree that experimental simplifications make debatable that robots are currently able to self-recognize in the same way that chimpanzees do. Therefore, instead of addressing the self-perception awareness level, we focused on body sensorimotor learning and integration and studying how these mechanisms aid in self-recognition and self/other distinction.

\begin{figure*}[t!]
	\centering
	\includegraphics[width=0.8\linewidth, height=180px]{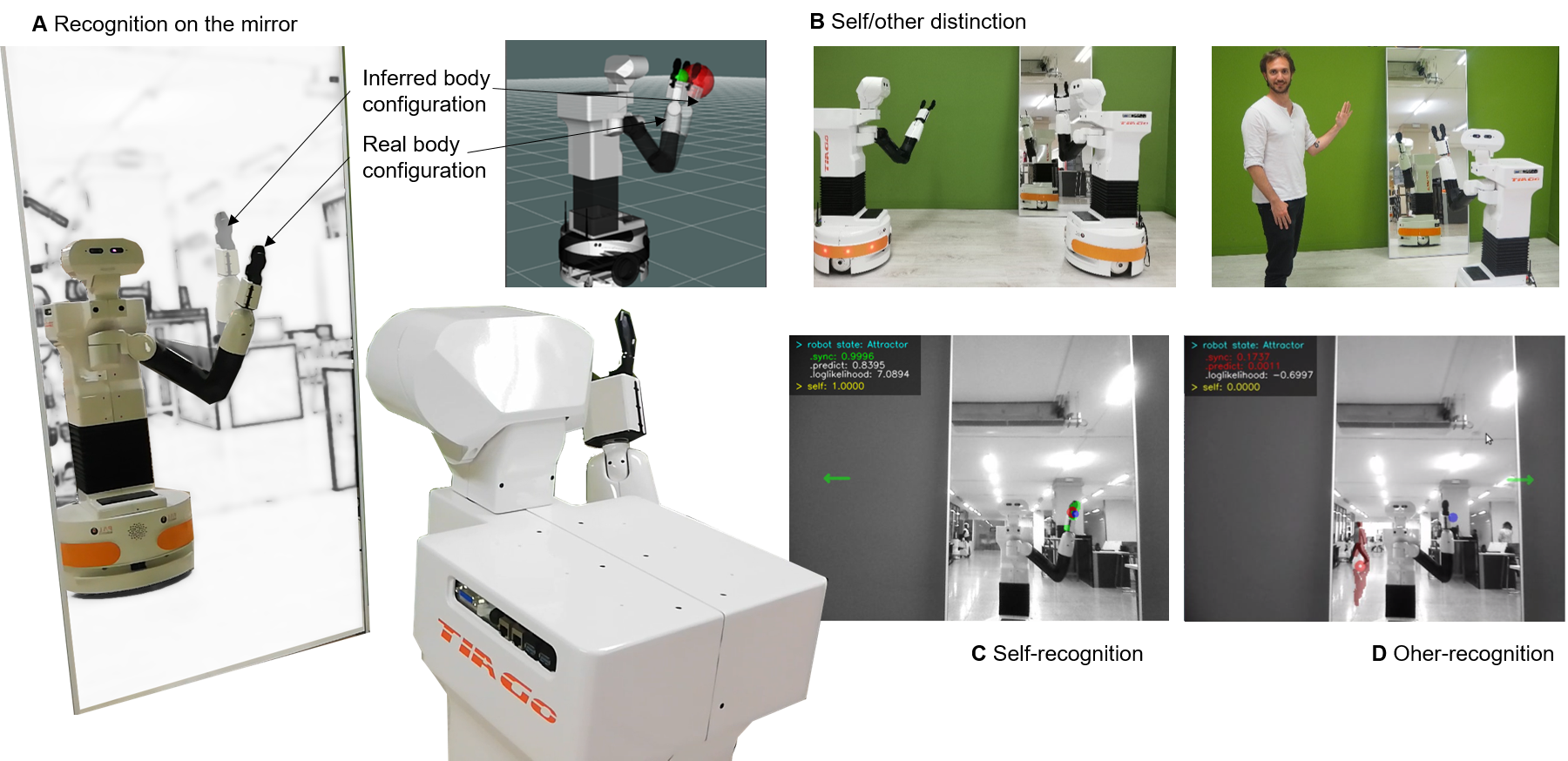}
	\caption{Self-recognition in the mirror and self/other distinction experiments}
	\label{fig:description}
\end{figure*}

\vspace{-2px}
\noindent\textbf{\\Relation to other approaches}

Artificial intelligence approaches for self-recognition are quite far from the biological plausibility \cite{lanillos2017enactive,georgie2019interdisciplinary}. However, some of them are already addressing relevant characteristics of self-perception such as timing, spatial correlations and contingency. 

On the one hand, we have methods that do not involve learning. In \cite{gold2009using} they used Bayesian inference to accumulate evidence. However, action and perception had to be binary variables and required continuous stable segmentation of the body as an entity, complicating its scalability to real scenarios. In \cite{lanillos2016yielding} this approach was extended to images and its relation with proprioception. The method depended on the skin accelerometers and the image grid subsampling. On the other hand, contingency is possible to be learnt with spiking neural networks \cite{pitti2017spatio} and self/other distinction can be developed by gradually increasing spatiotemporal resolution \cite{nagai2011emergence}.

Our approach involves both inference and learning and casts self/other distinction and self-recognition to free energy minimization (i.e. variational inference), in order to solve the scalability and partial information sensitivity limitations of previous approaches.


\noindent\textbf{\\Predictive coding approach}

Under the predictive coding \cite{rao1999predictive} approach, the robot is able to learn (approximate) generative models of the world, especially the effects of the body in the world. These prediction errors are used for inferring the body state but they are also an indirect measure to distinguish the robot generated actions.

Defining the brain as an inference machine \cite{knill2004bayesian} that is able to interpret the world from partial information \cite{Helmholtz1867, dayan1995helmholtz}, body perception is presented here as an unconscious process that adapts empirical information to current learned models. Based on the \textit{active inference} paradigm \cite{friston2010action, oliver2019active}, we formalized body perception and action as an optimization problem that minimizes the prediction error, i.e., the error between the expected sensory effect and the observed one~\cite{friston2010free}. According to active inference, adaptation can be performed by changing the interpretation or belief or by acting to modify the world according to our expectations. Thus, we deployed this adaptation mechanism on a humanoid robot and studied the perceptual response regarding self-recognition when executing movements in front of a mirror or in front of other agents.

\noindent\textbf{\\Contribution and organization}

This work presents a novel active inference computational model that combines free energy optimization from \cite{oliver2019active} with function learning from \cite{lanillos2018adaptive}. The learning was performed by means of a Mixed Density Networks, MDN \cite{bishop1994mixture}, to provide scalability and interpolation. The algorithm enables a robot to recognize itself in the mirror using non-appearance body cues and to distinguish its body from other entities. Our approach, as it incorporates learning, is robust to variations in the robot position and the mirror location and angle. On top of this model, we were able to compute the probability that our robot with a given set of parameters would generate a particular observed sensor value. The accumulation of this evidence yielded to self-recognize itself in the mirror and perform self/other distinction.

First, Section \ref{sec:methods} describes the experimental setup and the computational model. Section \ref{sec:results} shows the results and Section \ref{sec:discussion} analyze the relevance of the results with respect to robotics and artificial intelligence, and discusses its limitations regarding animal self-recognition.



\section{Methods}
\label{sec:methods}

\subsection{Experimental design}
\label{sec:methods:experiment}
\begin{figure}[hbtp!]
	\centering
	\includegraphics[width=0.58\linewidth, height=115px]{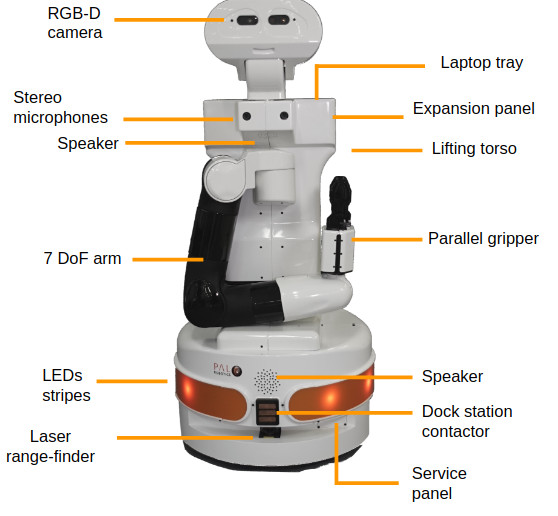}
	\caption{TIAGo robot used for the experiments}
	\label{fig:TIAGo_parts}
\end{figure}

Figure \ref{fig:description} shows the humanoid robot TIAGo \cite{pages2016} and the three different scenarios tested: in front of the mirror, robot-robot interaction and human-robot differentiation. TIAGo is a mobile manipulator by PAL Robotics that has been designed in a modular way. It is composed of a differential drive base; a lifting torso; a 7 DoF arm and a pan-tilt head. The robot has a laser rangefinder on the base and a RGBD camera on the head. For the experiments, we only used the monocular RGB camera as 3D computations are distorted in the mirror setting. In terms of manipulation capabilities, it has a  7 DoF arm and the lifting torso prismatic joint, being able to reach objects on the floor up to objects on shelves at around 1.75 m. For the self/other distinction experiment, only the arm will be used. Moreover, the robot has stereo microphones speakers and professional text-to-speech software. Figure \ref{fig:TIAGo_parts} shows the main modular components of the robot used.

\vspace{-0.6cm}

\subsection{Overall algorithm}
The algorithm \ref{alg:SO} summarizes the whole process computed in the robot for self-recognition and self/other differentiation. First, the visual input of the system is segmented using optical flow to extract the visual 2D centroid of any visual blob and the histogram of visual movement directions. This data will be used as the visual input $s_v$ and for computing the probability of being a contingent event respectively.
The predicted sensor values are computed by means of an MDN and its partial derivatives with respect to the latent space (i.e. joint angles). Then the error between the predicted and the current sensor input (e.g. visual error $e_v$) is computed and added into the differential equation weighted by its relevancy (i.e. variance $\Sigma_v$). The goal attractor dynamics are also computed for the left and the right condition. Finally, the probability of being generating that sensor input is computed.

If the error between the prediction and the observed values is too high (outlier), a learning process starts, where the robot retrain the MDN with the new input. The outcome depends on three conditions. (1) When there are no contingent effects the robot is not able to learn anything and therefore it is not itself. (2) When it is able to learn an effect but then the model cannot properly approximate the sensory effect then it is not itself either. Finally, (3) if it is able to learn the effect and the error is enough small then it will accumulate evidence until it will properly answer: \textit{it is me}.

\small
\begin{algorithm}[h!]
\caption{Self/other distinction algorithm}
\label{alg:SO}
\begin{algorithmic}[1]
\Require $\Sigma_p, \Sigma_v, \Sigma_\mu, \Delta_t$ LOGMIN, LOGMAX 
\State $\mu  \gets \text{Initial joints angle estimation}$
\State $\mu',v,\mathcal{L}  \gets \mathbf{0}$ 
\While{$\neg$ recognized($p_{self}$)}
    
    \State $h \gets of(I), I \gets \text{camera}$\Comment{Optical flow histogram}
    
    \State $p(cont) \gets sigmoid( \text{BCNN}.forward(a,h))$\Comment{Sec. \ref{sec:sync}}
    \State $g,\mu^*, \Sigma^* \gets  \text{MDN}.forward(\mu)$ \Comment{Predictor Sec. \ref{sec:MDN}}
    \State $\partial_\mu g \gets  \text{MDN}.backward(\mu^*, \Sigma^*)$ \Comment{Sec. \ref{sec:MDN}}
    
    \State $\dot{\mu} \gets \mathbf{0}$ \Comment{Active Inference. Sec \ref{sec:ai}}
        \If{$\exists s_p $} \Comment{Proprioception}
        \State $e_p = (s_p - \mu)$ 
        \State $\dot{\mu} = \dot{\mu} + e_p/\Sigma_p$
        \State $\dot{a} = \dot{a} - (e_p/\Sigma_p)\Delta_t$
    \EndIf
    \If{$\exists s_v $} \Comment{Vision}
        \State $e_v = (s_v - g)$ 
        \State $\dot{\mu} = \dot{\mu} + \partial_\mu g^T e_v/\Sigma_v$
        \State $\dot{a} = \dot{a} - (\partial_\mu g^T e_v/\Sigma_v)\Delta_t$
    \EndIf
    \If{$\exists s_\rho $} \Comment{Goal attractor $\rho$ dynamics}
        \State $e_f = (\mu' - f(\mu,\rho))$ 
        \State $\dot{\mu} = \dot{\mu} + \partial_\mu f^T e_f/\Sigma_\mu$
    \EndIf
    \State $\mu = \mu + \Delta_t \dot{\mu}$ \Comment{1st order Euler integration}
    
    \State $a = a + \Delta_t \dot{a}$
     \If{$\exists s_v $} \Comment{Self-recognition Sec. \ref{sec:pself}}
        \State $\mathcal{L}_i = -0.5(e_v^T \Sigma_v^{-1} e_v + \ln \det{\Sigma_v} + n\ln2\pi)$
        \State $\mathcal{L} = \mathcal{L} + \mathcal{L}_i + \ln p(cont)$ 
        \State $p_{norm}\!=\!\! \left(\exp(\mathcal{L})\!\!-\!\!  \text{LOGMIN}\right)/(\text{LOGMAX} - \text{LOGMIN})$
        \State $p_{self} = p_{self} + K(p_{norm} -p_{self})$
    \EndIf
    \If{$\text{outlier}$} 
        \While{$i < N_s \wedge \neg \text{timeout}$}
            \If{$p(cont) > \delta$} \Comment{Learning Sec. \ref{sec:sync}}
                \State $Qx.append(s_p)$; $Qy.append(s_v)$
            \EndIf
            \If{$i > N_s$}
                \State $\text{MDN}.train(Qx,Qy)$
            \EndIf
        \EndWhile
    \EndIf
\EndWhile
\end{algorithmic}
\vspace{-2px}
\end{algorithm}
\normalsize
\vspace{-6px}

\subsection{Detailed computational model}
\label{sec:methods:model}

\begin{figure}[hbtp!]
	\centering
	\includegraphics[width=0.95\linewidth]{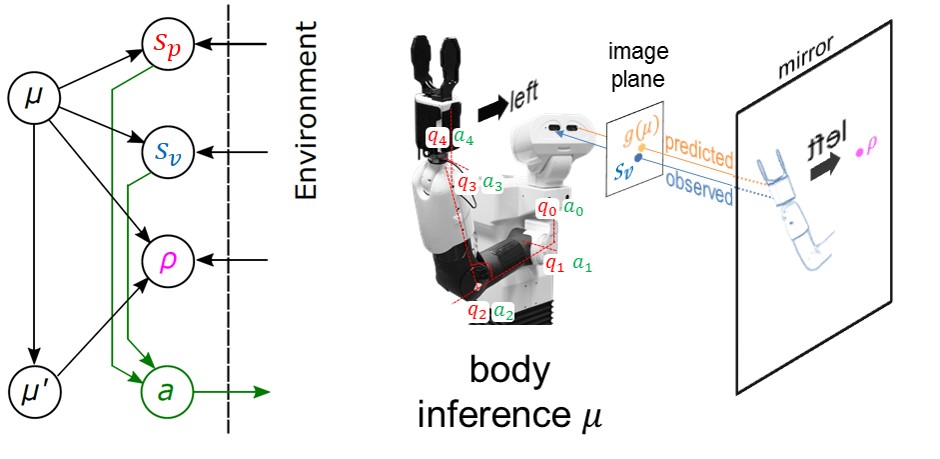}
	\caption{Body perception model and notation. (Left) Generative model of the robot perception and action. (Right) Mirror and robot setup. $q$ are the joint measurements and $\mu,\mu'$ is the inferred angle of the joints and its first-order dynamics. $v$ is the visual location of the hand and $\rho$ the goal. $a_i$ is the action exerted in joint $i$.}
	\label{fig:model}
\end{figure}

We describe in detail the body perception and action model and then the method for self/other distinction. We define body perception as inferring joint angles of the robot arm $\mu$ by means of different sensory modalities. The robot approximates the belief of his body by means of minimizing the free energy \cite{friston2010free}, computing the most plausible solution of his arm location using the error between the predicted sensory input and the observed one. Mathematically, perception and action are interpreted by the following two equations adapted from the free energy principle approach model proposed in \cite{friston2010action}:

\begin{align}
    &\text{Perception} \quad \mu = \arg\min_{\mu} F \rightarrow \dot{\mu} = \mu' - \frac{\partial F(\mu, \rho)}{\partial \mu} \label{eq:overallperception}\\
    &\text{Action} \quad \quad a = \arg\min_{a} F \rightarrow  \dot{a} =- \frac{\partial F(\mu, \rho)}{\partial a}
    \label{eq:overallaction}
\end{align}

\subsubsection{Active inference model}
\label{sec:ai}
We adapted the free energy model presented in \cite{oliver2019active} to the TIAGo robot. Using joint encoders $s_p$ and visual information $s_v$, the graphical representation of the inference process is described in Fig. \ref{fig:model}. The agent encodes its belief of its body $\mu$ with the following density \cite{buckley2017free}: 
\begin{align}
p(\mu,s, \rho) = p(s_p|\mu) p(s_v|\mu) p(\mu'|\mu,\rho)
\end{align}
Where the first two likelihoods are the sensor modality contributions (assumed independent between each other) and the third term is the prior density for the first-order dynamics, which also depends on the causal variables $\rho$. We will control the robot in joint velocities by defining actions $a$ as the velocity of each joint, thus only first-order dynamics are needed.

Under the mean-field Laplace approximation, the free energy bound can be approximated as\footnote{In some formulations of the free energy the sensor inputs $s$ and other causal variables are the described by the same variable. However, in this paper, we split them for clarity purposes.}:
\begin{align}
F \approx -\ln p(\{s,\rho\},\mu) - \left[ \frac{1}{2} \ln | \Sigma | + n \ln 2\pi \right]
\end{align}
where $n$ is the size of $\mu$. In order to solve Eq. \ref{eq:overallperception} and \ref{eq:overallaction}, we computed the gradient of $F$ with respect to the latent variables $\mu, a$.

\begin{align}
\label{eq:mu}
\frac{\partial F}{\partial\mu} = \frac{(s_p-\mu)}{\Sigma_p} &+ \frac{\partial g(\mu)^T}{\partial \mu}\frac{(s_v-g(\mu))}{\Sigma_v} +\\&+ \frac{\partial f(\mu,\rho)^T}{\partial \mu}\frac{(\mu'-f(\mu,\rho))}{\Sigma_\mu}
\end{align}

\begin{align}
\label{eq:v}
\frac{\partial F}{\partial a} = \frac{\partial s_p}{\partial a}^T \frac{(s_p-\mu)}{\Sigma_p} + \frac{\partial s_v}{\partial a}^T \frac{(s_v-g(\mu))}{\Sigma_v}
\end{align}
where $g_v(\mu)$ is the visual generative model that is learnt (Sec. \ref{sec:MDN}) and $f(\mu, \rho)$ is the model dynamics of a perceptual attractor \cite{oliver2019active}.

\subsubsection{Generative models learning}
\label{sec:MDN}
To solve Eq. \ref{eq:mu} and \ref{eq:v} the generative models for the sensor $g(\mu)$ and its partial derivative $\partial g/\partial \mu$ should be known. Conversely to previous works on active inference we enabled the robot to learn those models taking inspiration from our latest contributions on body perception with function learning~\cite{lanillos2018adaptive,lanillos2018active}. However, in this case, we used a probabilistic neural network model to provide scalability and interpolation. The visual generative model is learnt by means of a Mixture Density Network (MDN) \cite{bishop1994mixture}. In this way, we encode not only the mapping but the uncertainty associated. In principle, the forward model of the robot end-effector with respect to the joint latent space can be learnt using just one Gaussian kernel. However, we allowed the robot to learn more complex visual patterns generated by the actions or due to the lens non-linearity. The same model can generalize to forward model learning using more complex probabilistic deep models \cite{chua2018deep} within the free energy framework. The predicted sensation is computed through the forward pass of the MDN and outputs the Gaussian mixture model parameters: $\Pi_{i}$ coefficients, $\overline{s}_{i}$ mean and $\Sigma_i$ variance. To compute the prediction of the sensation given the state $s = g(\mu)$ we select the maximum probable kernel and get the mean and the variance as follows: 

\begin{align}
&i^* = \arg\max_i \quad \text{softmax}(W_{\Pi_i}^T h(\mu) + b_\Pi) \\
&g(\mu) = W_{\overline{s}}^T h(\mu) + b_{\overline{s}}
\end{align}
where $h(\mu)$ is the hidden layer output and $W,b$ are the network weights and bias respectively. Although for this specific visual location learning only one hidden layer is sufficient, more layers can be added for more complex non-linear function learning.

\begin{figure}[hbtp!]
	\centering
	\includegraphics[width=0.99\linewidth]{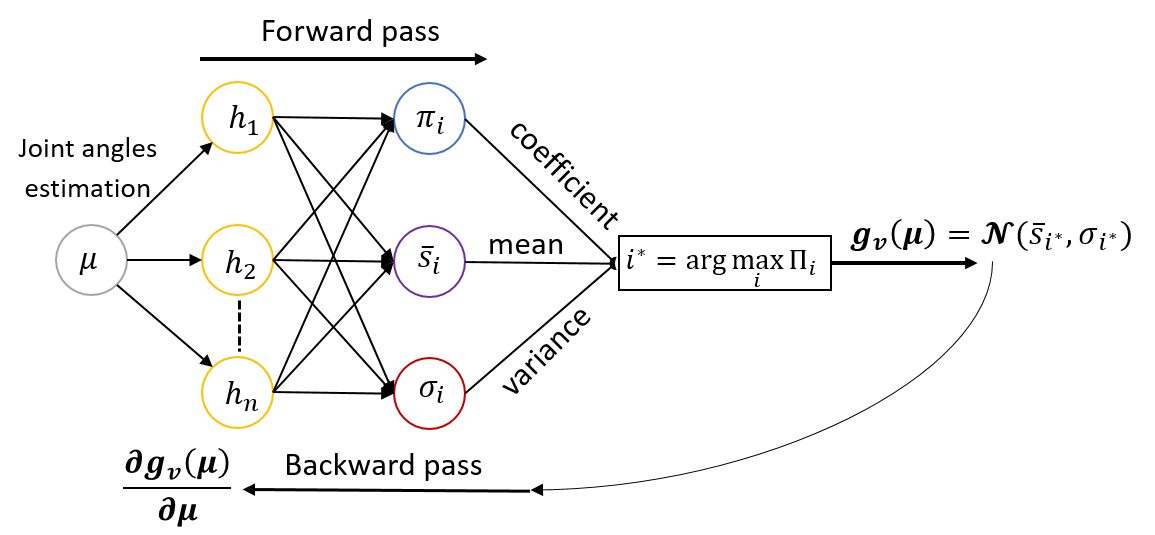}
	\caption{Forward model learning with MDN. The forward pass gives the prediction and the backward pass gives the partial derivative with respect to the body configuration $\mu$.}
	\label{fig:mdn}
\end{figure}

\begin{figure}[hbtp!]
	\centering
	\subfigure[Training samples and $g(\mu)$]{\includegraphics[width=0.45\linewidth]{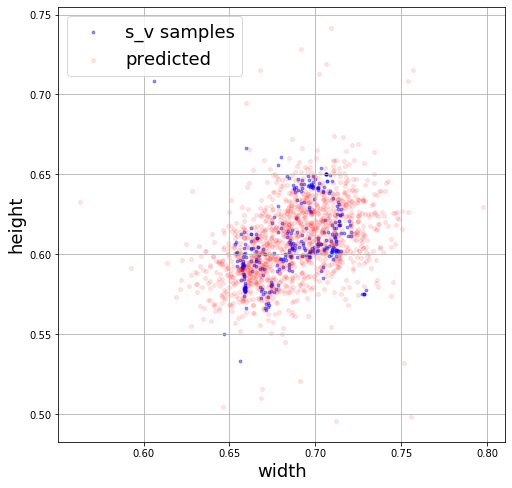}}
	\subfigure[MDN training loss]{\includegraphics[width=0.45\linewidth]{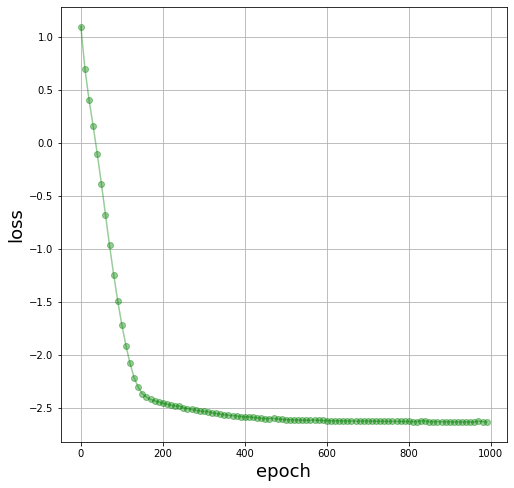}}
	\caption{Learned mapping with the MDN from the joints positions to the 2D visual point (end-effector) computed through optical flow. Visual samples (blue) used for training and $g(\mu)$ predictor learned (red). Axis ranges that correspond to the image dimensions have been normalized between 0 and 1 dividing by the width (640px) and the height (480px) respectively.}
	\label{fig:mdn:train}
\end{figure}

Figure \ref{fig:mdn} describes the MDN to obtain the forward and partial derivatives through the backward pass. Figure \ref{fig:mdn:train} shows an instance of the visual forward model training for the robot in front of the mirror. The negative log-likelihood converges rapidly in 400 epochs. The blue dots correspond to the visual samples acquired when the robot moves the arm and the red dots are generated with the learnt predictor $g(\mu)$ for random joint angles within the set of the waving left and right movement. The interpolation power of the network shown is really crucial for the stability of free energy optimization.

The inverse mapping or partial derivative with respect to the latent variable is computed through the backward pass of the MDN of the most plausible Gaussian kernel $i^*$.

\begin{align}
\frac{\partial g(\mu)}{\partial \mu} = P_{i^*} \frac{(\overline{s}_{i^*} -s_v)}{\sigma_{i^*}^2} \frac{\partial z_{i^*}}{\partial \mu}
\end{align}
where $z_i$ is the neuron output, $\overline{s}_i$ is the mean value output of the selected kernel, $s_v$ is the current observation of the end-effector in the visual field and $P_{i^*}= \Pi_{i^*}\mathcal{N}(\overline{s}_{i^*},\sigma_{i^*}^2) / \sum_j \Pi_j\mathcal{N}(\overline{s}_{j},\sigma_{j}^2)$ - See \cite{bishop1994mixture}.

In order to train the neural network we use as the input the joints estimated angles $\mu$ and as the output the 2D centroid location of the biggest moving optical flow blob. The points that are contingent (see Sec. \ref{sec:sync}) are buffered into a batch that is then used to modify network weights through backpropagation \cite{kingma2014adam}. The loss function used is the standard negative log-likelihood of the Gaussian mixture model: $\mathcal{L} = -\ln \sum_i \Pi_i\mathcal{N}(\overline{s}_i, \sigma_i^2).$

\subsubsection{Contingency learning}
\label{sec:sync}
\begin{figure}[hbtp!]
	\centering
	\includegraphics[width=0.99\linewidth]{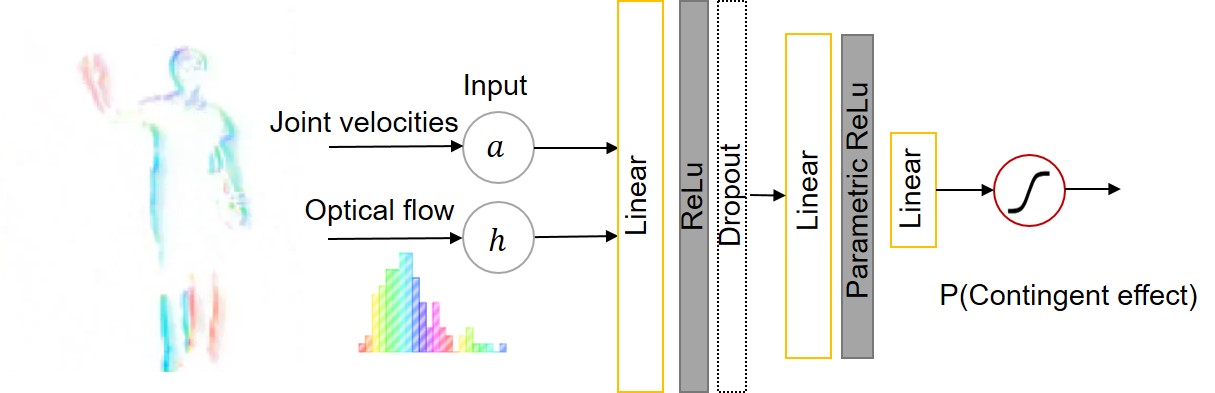}
	\caption{Contingent sensory event classifier. Given the histogram of the optical flow velocities of the visual field (left), the joint states and the velocities being exerted, the deep binary classifier computes whether the effects are contingent for the left and right action.}
	\label{fig:contingency}
\end{figure}

We assume that the robot has a system that is able to classify whether it is a contingent effect or not. This prior knowledge can be justified by stating that the robot has learnt the potential effect of moving the hand left and right in front of him. In essence, the robot has never looked in a mirror but has performed actions with the arm that result in the following contingent effect learning: when the robot moves the hand to the left produces a change in the visual field with a specific direction. We computed the effect in the visual field given the action (i.e. velocity) by means of an optical flow algorithm \cite{simoncelliBOF}. To learn to classify this effect we used a deep learning classifier with input the state of the robot (i.e. estimated joint states $\mu$) and the velocity being exerted (i.e. $a$) and the output the histogram of the optical flow velocity directions. Figure \ref{fig:contingency} shows the input, the output and the architecture of the neural network.

\begin{figure}[hbtp!]
	\centering
	\subfigure[Joint velocities and OF histogram]{\includegraphics[width=0.32\linewidth]{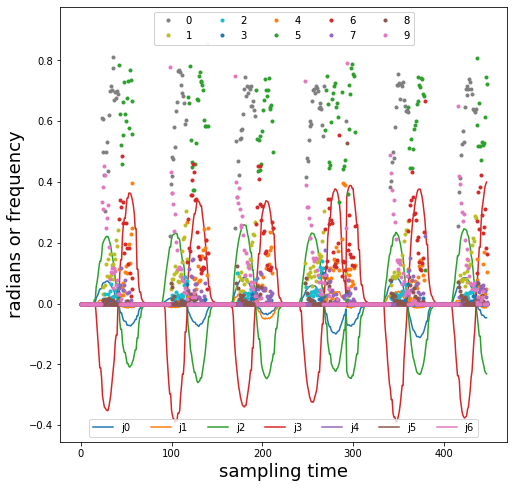}}
	\subfigure[Training loss]{\includegraphics[width=0.32\linewidth]{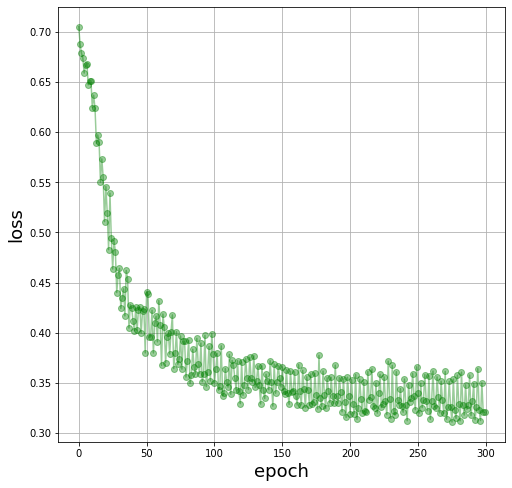}}
	\subfigure[Classification with test data]{\includegraphics[width=0.32
	\linewidth]{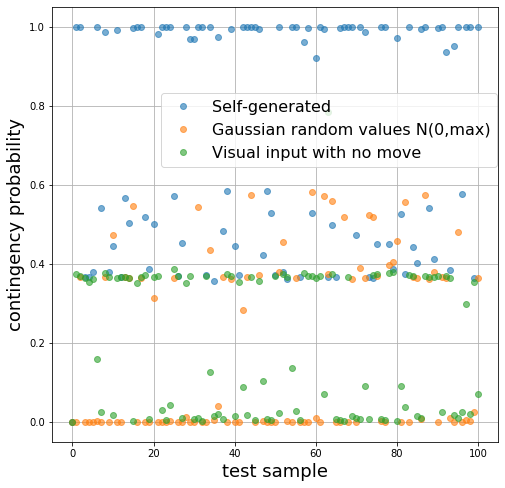} \label{fig:bccn:train:c}}
	\caption{Contingency classifier learning. The input are the 7 joint velocities (lines) and the optical flow histogram with 10 bins (dots). The output is the probability of a image velocity vector generated by an action on the joints.}
	\label{fig:bccn:train}
\end{figure}

Figure \ref{fig:bccn:train} shows the input data, the learning curve and the classification of samples of the test set. The joint velocities (coloured lines) for the hand wave movement show the profiles of left and right movements. The associated optical flow histogram (coloured dots) varies depending on the movement. The orientations are split into 10 bins represented with different colours and the y-axis is the normalized frequency. The loss function used is the binary cross-entropy with logistic output. Then we used a sigmoid function to compute the probability. The optimizer was Adam with 0.01 learning rate. It is important to highlight that we had to include the non-movement case to deal with the sensor noise. We tested, in Fig. \ref{fig:bccn:train:c}, 100 random points from the test set where we included: self-generated actions, Gaussian random noise and non-movement with visual input. The contingency probability (y-axis) is shown for all those cases. Furthermore, some random samples have some probability of being classified as self-generated.

\vspace{-6mm}
\subsubsection{Self-recognition: Is that me?}
\label{sec:pself}
In order to provide non-appearance self-recognition, we transform the question of \textit{is that me?} into \textit{did I generate those sensors values?} The robot accumulates evidence of generating the sensed effects in the world. This is computed through the marginal likelihood or sensory surprise. Highly surprising events (effects in the visual field) drastically drop the probability of being self. Conversely, in order to increase the probability of being self, the effect should continuously match the expected value generated by the learnt forward model.

We compute the probability that our model with a given set of parameters would generate a particular observed cue in the visual field as:
\begin{align}
\mathcal{L}_i &= -\frac{1}{2}(e_v^T \Sigma_v^{-1} e_v + \ln \det{\Sigma_v} + n\ln2\pi)\\
\mathcal{L} &= \mathcal{L} + \mathcal{L}_i + \ln p(contingent)
\end{align}
where $p(contingent)$ is computed with the deep net classifier of Sec. \ref{sec:sync}.
Inspired by \cite{kahl2018predictive}, the probability of being itself is computed by normalization and smoothing as follows:
\small
\begin{align}
p_{norm} &= \left(\exp(\mathcal{L}) - \text{LOGMIN}\right) /(\text{LOGMAX} - \text{LOGMIN})\\
p_{self} &= p_{self} + K(p_{norm} -p_{self})
\end{align}
\normalsize
where $K$ is a smoothing constant working as a Kalman gain.

\section{Experimental results}
\label{sec:results}
Two experiments were designed: self-recognition in the mirror, depicted in Fig. \ref{fig:description}A,  and self/other distinction with other similar robot and a human, represented in Fig. \ref{fig:description}B. A video describing the experiments is here: \url{https://youtu.be/3l9N972xjD8}. The parameters of the MDN network were fixed for all experiments: $\mu \in \mathbb{R}^7$, one hidden layer with 20 units, two-dimensional output $s_i \in \mathbb{R}^2$ (i.e. hand 2D position in the image). The kernels were Gaussian distributions $\mathcal{N}$ with mean $\overline{s}_i$ and variance $\sigma_i$. We assumed that the output variables (2D location in the image) have the same noise distribution, thus we only used one variance parameter for each Gaussian kernel. We fixed the number of Gaussian mixtures to four ($i \in (0,4)$). The optimizer used was Adam \cite{kingma2014adam} and the training used batches of 200 samples and 1000 epochs.

\subsection{Self-recognition in the mirror}
The robot was placed in 10 random positions in front of a body size mirror (Fig.~\ref{fig:description}A) and we ran the experiment 10 times for each position. The arm was always visible in the mirror and we did not include any prior information about the mirror. All experiments were successfully classified as itself.

The procedure went as follows. The robot moved the hand left and right by setting an attractor of the end-effector in the visual field. In other words, the desired location in the image changed from left to right of the image centre. The attractors were active for 2 seconds each. Then, the delay between initiating the left and right waving changed every time randomly in the range of 3 seconds. This enforced that actions were not periodic. Finally, the training finished if the number of samples was more than 200 or it reached 20 seconds. 

\begin{figure*}[hbtp!]
	\centering
	\subfigure[self-recognition in the mirror\label{fig:results:a}]{%
       \includegraphics[width=0.43\linewidth, height=145px]{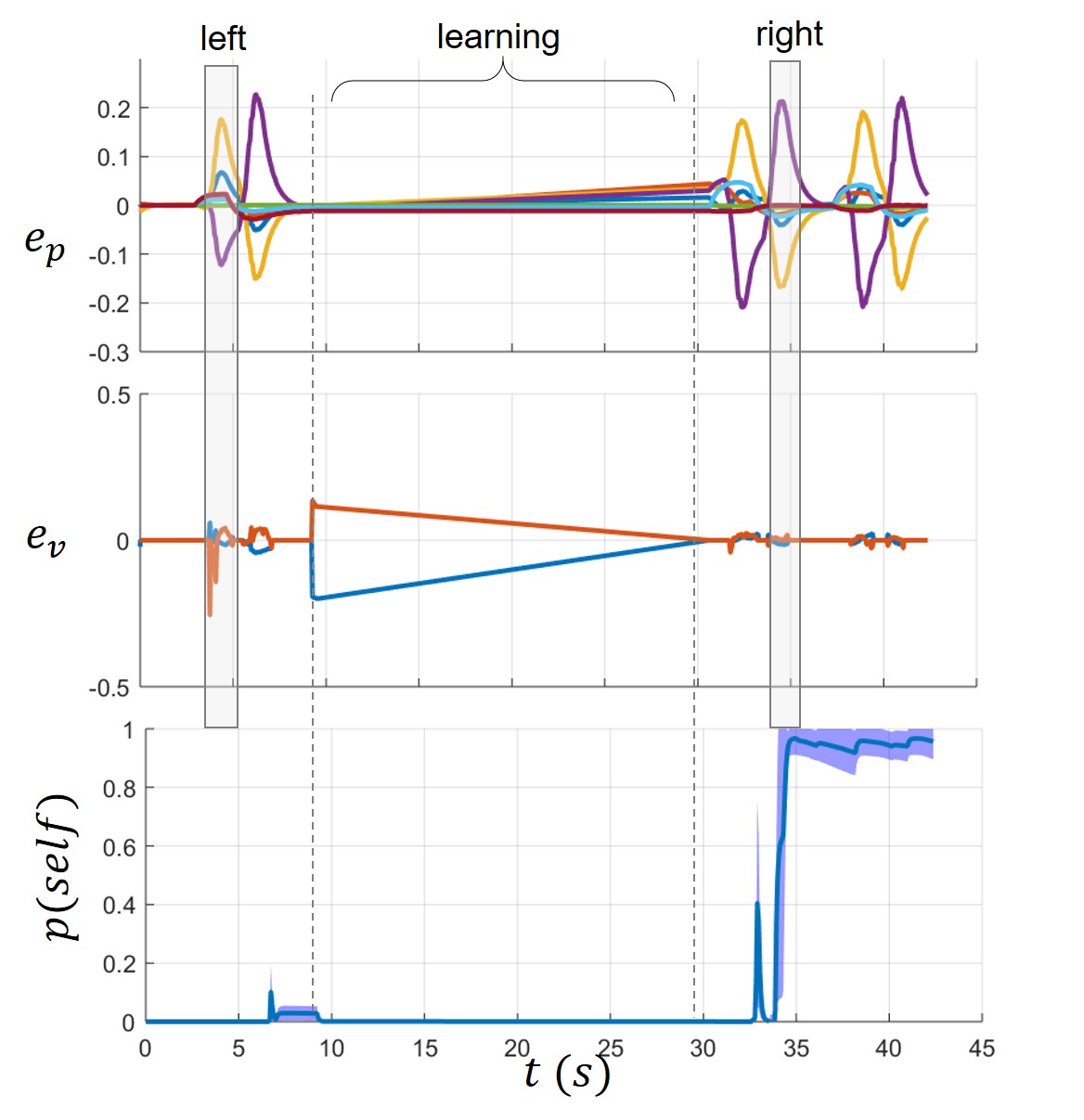}}
    \subfigure[self/other distinction\label{fig:results:b}]{%
        \includegraphics[width=0.43\linewidth, height=145px]{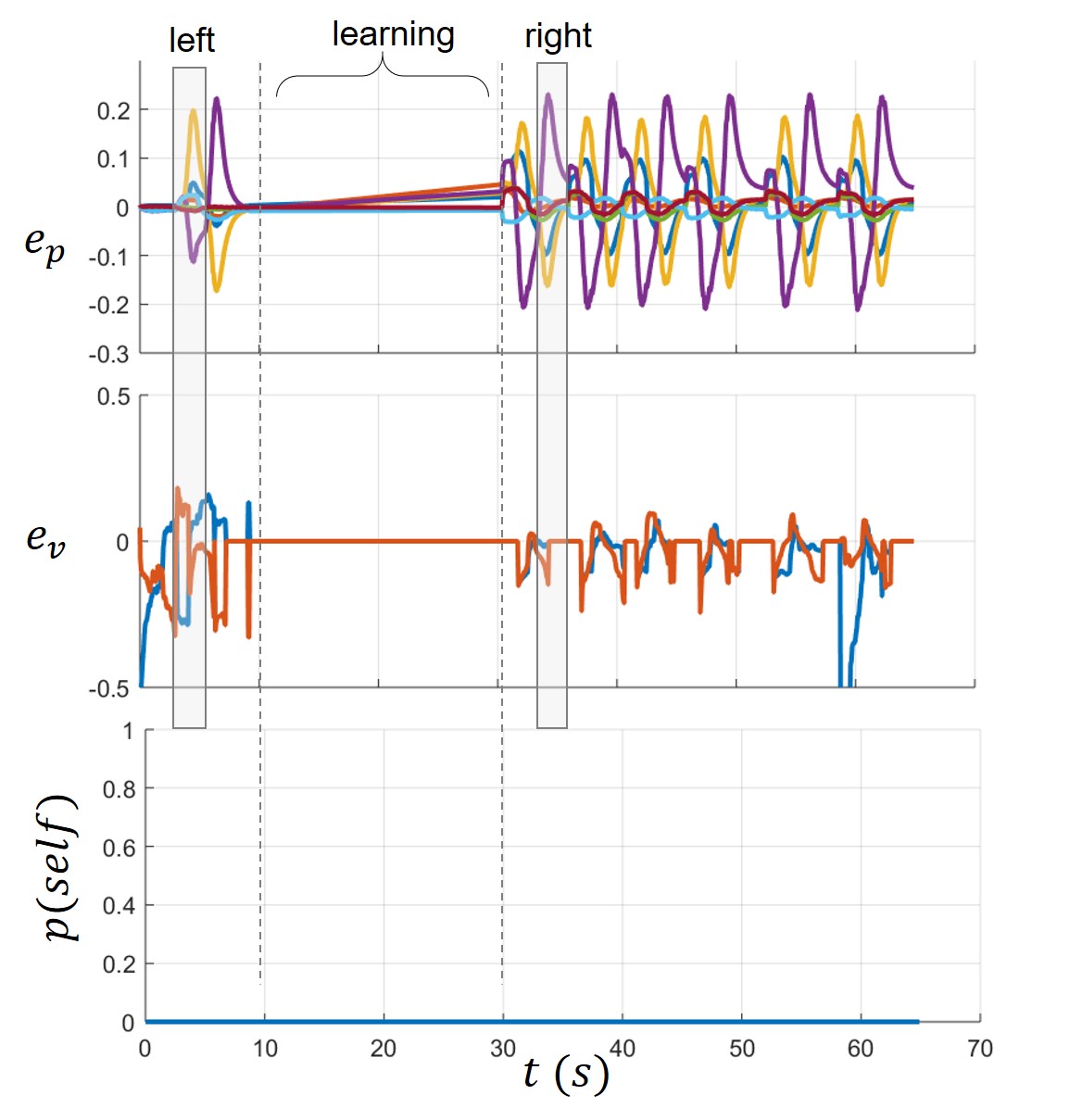}}
	\caption{Self-recognition and self/other distinction experiments variable dynamics.$e_p$ and $e_v$ are the proprioceptive and visual errors. During learning, the active inference update is disabled. The probability of being self, $p(self)$, rises when the predicted visual effect is contingent and matches the observation.}
	\label{fig:results}
\end{figure*}

 Figure \ref{fig:results:a} shows the prediction error dynamics for proprioception $e_p$ and visual $s_v$ cues for one trial\footnote{Other trials have similar profiles but with different delays. Thus, they don't bring any further information.}. The bottom plot shows the mean and standard deviation of the probability of being self in all trials with no prior training with the mirror. The first time that the robot looks at the mirror and starts moving the arm, the robot is not able to predict its hand position producing a big prediction error. Before the training stage, the robot was not able to recognize itself. Thus, it starts learning the relation between its body state and the visual input by means of the MDN (Sec. \ref{sec:MDN}). After learning for 10 seconds, the robot continues moving the hand left and right until it decides whether it is him or not. After learning the forward model the error in the visual sensation drops allowing the system to accumulate evidence of being itself (the model is correctly generating the visual sensations). Still, there are small prediction errors related to the adaptation of the current learnt model and real observations. Errors in proprioception are generated by the perceptual attractor created when it has the goal to move to the left or to the right. Any disturbance in the expected visual sensation (surprise) dropped the probability of being self.

\subsection{Self/other distinction}
For the self/other experiment, we performed the same amount of trials. We faced two similar robots doing the same movements but due to the randomness of the delays between actions they were asynchronous (Fig.~\ref{fig:description}B). As well as in the previous experiment, figure \ref{fig:results:b} shows the error in proprioceptive and visual cues of one of the trials. The bottom plot of Fig. \ref{fig:results:b} shows the probability of being self for all trials. In this experiment, the robot could not learn anything as there was no causality between the actions (joint velocities) and the observed sensory information. This yielded into the big visual prediction errors after the training. Therefore the robot inferred that it is not itself.

We further tested the model with an experimenter in front (Fig.~ \ref{fig:description}B) and we observed that even when trying to perform the same movements the robot was able to properly distinguish between self-generated and other generated visual cues. In all cases, the robot correctly identified the agent in front as not itself. However, as expected, in the case of perfect synchronization the robot will identify it as itself. From the cognition point of view interpreting another robot as itself seems wrong, but also humans will perceive agency when seeing an arm that moves at every instant with correlation with their commands to the muscles.

\section{Machines able to recognize themselves}
\label{sec:discussion}

We pointed out that unconscious body perception, action and learning is underneath self-recognition. Experiments showed that contingent non-appearance cues can be learned through interaction and be used to perform self/other distinction. The proposed algorithm enabled the robot to robustly identify itself in the mirror just by learning the forward model, i.e., the expected visual input given the inferred body state. As long as the arm appeared in the mirror, the initial conditions where irrelevant. This is in line with the experiments with non-human primates that needed learning with the mirror before achieving the self-recognition ability.

The robot was also able to differentiate itself from other entities. The algorithm was sensitive to any unexpected sensations as we are using the minimization of surprise as the key mechanism for the differentiation. Even when trying to deceive the algorithm (e.g. robot-human experiment), any small delay or difference dropped down the probability of being itself. Interestingly, the variance parameters (relevance of the sensory modality for inferring the body) was crucial for dealing with the robot own noise but differentiating from other entities cues. We further observed an interesting side effect of using the action to minimize the prediction error. The human or robot hand in the self/other distinction experiment acted sometimes as an attractor for the body inference, generating mirroring actions similar to imitation.

In summary, with these experiments, we are showing that self/other distinction and self-recognition is possible to be implemented in artificial agents. Although it will not be a conscious cognitive process as described in humans or elephants, it will have some common ground in the sensorimotor action-perception level.

\noindent\textbf{\\Double comparator model}

Within the self-perception cognitive science literature it is common knowledge the comparator model \cite{david2008sense}. This approach determines that humans are aware of producing an effect in the world through their actions (i.e. agency \cite{hommel2015action}) by means of the prediction error congruence. Some researchers criticized the simplistic vision of the model and proposed that contingency detection and causal inference should be also involved as processes on top of congruence \cite{zaadnoordijk2019match}. This is similar to our model. We use both prediction error and contingency learning. In essence, we propose the ``double comparator" model inspired by the work of \cite{wegner2017illusion}. The first comparator is the prediction error mechanism that approximate the models learnt to reality for improving interpretation. The second comparator takes into account other processes like spatio-temporal contingency and performs the proper distinction. In our model, we engineered this abstract level using a probabilistic approach, i.e., by calculating the marginal likelihood to accumulate temporal evidence and then computing the probability of generating that observed effect in the sensory space. 

\noindent\textbf{\\Towards passing the original mirror test}

We showed how non-appearance cues can be used for self-recognition. However, robots are still not able to pass the original self-recognition test like animals do~\cite{anderson2011primates}. For instance, the robot here did not learn the mapping between the tactile and the visual reflection to produce the motor response. In order to fully provide robots with a full-fledged self-recognition similar to humans, there are a set of technical challenges that should be further addressed:
\vspace{-2px}
\begin{itemize}
    \item\textit{Complex visual segmentation.} The algorithm was designed for only one moving blob segmentation using optical flow. Complex inputs with multiple causal sources need improved segmentation.
    \item\textit{Large scale input data.} Forward models learning of the end-effector location in the image should be scaled to real sensory input data, such as images. 
    \item\textit{Whole body interactions.} We used the arm for the experiments as a proof-of-concept. The same approach can be extended to whole-body self/other distinction.
    \item\textit{Other cues.} The proprioceptive-visual tandem is just one of the sensory inputs involved in self-recognition. We did not model appearance, tactile or auditory cues. We considered movement contingency as the first level of self-distinction but then an appearance model similar to the body image should be included.
    \item\textit{Biological plausibility.} Although the mathematical abstraction of body perception has strong neuroscience-inspired notions, using the marginal likelihood for accumulating evidence does not have a direct relation with the mechanism observed in the brain. However, it has roots into the model evidence approach under the Bayesian brain assumption.
\end{itemize}

\section{Conclusion}
\label{sec:conclusions}
We proposed an algorithm for self/other distinction using non-appearance sensory cues by combining free energy optimization with neural network learning. The approach showed the importance of not only learning the models of the body but also approximating those models to the observed reality. After learning the humanoid robot was able to accumulate evidence during the generated movements to discern whether it was itself or other entity in different scenarios. This reliability is due to the explicit handling of the sensorimotor uncertainty in the proposed model, while still performing robust self-differentiation. When the expected sensation was similar to the sensor values, i.e. the prediction error was low, the evidence of being itself was accumulated. Conversely, any prediction error dropped down the probability of being itself yielding to self/other distinction. This work stresses that self-recognition and self/other distinction grounds on the body perception and action model and that it is possible to give artificial agents, such as robots, these capabilities. Further work will focus on scaling to raw images \cite{sancaktar2019end} and including appearance cues.
\vspace{-4px}
\ack This work was supported by the MSCA SELFCEPTION project (www.selfception.eu) EU H2020 grant no. 741941 and PAL robotics.
\vspace{-0.3cm}

\bibliography{pl,selfception}

\end{document}